\documentclass[manuscript]{acmart}
\AtBeginDocument{%
  \providecommand\BibTeX{{%
    \normalfont B\kern-0.5em{\scshape i\kern-0.25em b}\kern-0.8em\TeX}}}

\setcopyright{acmlicensed}
\copyrightyear{2024}
\acmYear{2024}
\acmDOI{XXXXXXX.XXXXXXX}

\acmConference[ICER '24]{Proceedings of the 20th ACM Conference on International Computing Education Research}{August 13--15, 2024}{Melbourne, Australia}
\acmISBN{978-1-4503-XXXX-X/18/06}

\usepackage{todonotes}
\usepackage{makecell}
\usepackage{multirow}
\usepackage{tabularx}
\usepackage{natbib}

\def\mfor{\textbf{Forming}}
\def\mdis{\textbf{Dislodging}}
\def\mass{\textbf{Assumption}}
\def\mloc{\textbf{Location}}
\def\mach{\textbf{Achievement}}
\def\mint{\textbf{Interruption}}
\def\mpro{\textbf{Progression}}
\def\mmis{\textbf{Mislead}}
\def\m-d{metacognitive difficulties}
\def\M-d{Metacognitive difficulties}
\begin{document}


%\title{Generative AI for Novice Programmers Considered Harmful}
%\title{``I got this!'': Illusions of Competence for Novice Programmers with Generative AI}
\title{The Widening Gap: The Benefits and Harms of Generative AI for Novice Programmers}

\author{James Prather}
\orcid{}
\affiliation{
  \institution{Abilene Christian University}
  \city{Abilene}
  \country{TX, USA}
}
\email{james.prather@acu.edu}

\author{Brent Reeves}
\orcid{}
\affiliation{
  \institution{Abilene Christian University}
  \city{Abilene}
  \country{TX, USA}
}
\email{brent.reeves@acu.edu}

\author{Juho Leinonen}
\orcid{0000-0001-6829-9449}
\affiliation{
  \institution{Aalto University}
  \city{Aalto}
  \country{Finland}
}
\email{juho.2.leinonen@aalto.fi}

\author{Stephen MacNeil}
\orcid{}
\affiliation{
  \institution{Temple University}
  \city{Philadelphia}
  \country{PA, USA}
}
\email{stephen.macneil@temple.edu}

\author{Arisoa S. Randrianasolo}
\orcid{}
\affiliation{
  \institution{Abilene Christian University}
  \city{Abilene}
  \country{TX, USA}
}
\email{sar04b@acu.edu}

\author{Brett Becker}
\orcid{}
\affiliation{
  \institution{University College Dublin}
  \city{Dublin}
  \country{Ireland}
}
\email{brett.becker@ucd.ie}

\author{Bailey Kimmel}
\orcid{}
\affiliation{
  \institution{Abilene Christian University}
  \city{Abilene}
  \country{TX, USA}
}
\email{blk20c@acu.edu}

\author{Jared Wright}
\orcid{}
\affiliation{
  \institution{Abilene Christian University}
  \city{Abilene}
  \country{TX, USA}
}
\email{jpw19b@acu.edu}

\author{Ben Briggs}
\orcid{}
\affiliation{
  \institution{Abilene Christian University}
  \city{Abilene}
  \country{TX, USA}
}
\email{bab19b@acu.edu}

\renewcommand{\shortauthors}{Prather et al.}

\begin{abstract}
Novice programmers often struggle through programming problem solving due to a lack of metacognitive awareness and strategies. Previous research has shown that novices can encounter multiple metacognitive difficulties while programming, such as forming incorrect conceptual models of the problem or having a false sense of progress after testing their solution. Novices are typically unaware of how these difficulties are hindering their progress. Meanwhile, many novices are now programming with generative AI (GenAI), which can provide complete solutions to most introductory programming problems, code suggestions, hints for next steps when stuck, and explain cryptic error messages. Its impact on novice metacognition has only started to be explored. Here we replicate a previous study that examined novice programming problem solving behavior and extend it by incorporating GenAI tools. Through 21 lab sessions consisting of participant observation, interview, and eye tracking, we explore how novices are coding with GenAI tools. Although 20 of 21 students completed the assigned programming problem, our findings show an unfortunate divide in the use of GenAI tools between students who accelerated and students who struggled. Students who accelerated were able to use GenAI to create code they already intended to make and were able to ignore unhelpful or incorrect inline code suggestions. But for students who struggled, our findings indicate that previously known metacognitive difficulties persist, and that GenAI unfortunately can compound them and even introduce new metacognitive difficulties. Furthermore, struggling students often expressed cognitive dissonance about their problem solving ability, thought they performed better than they did, and finished with an illusion of competence. Based on our observations from both groups, we propose ways to scaffold the novice GenAI experience and make suggestions for future work.

%We argue that these findings indicate a need for GenAI scaffolding through new tools and pedagogies to combat this concerning problem.
\end{abstract}






\maketitle

%=========================
%
%   Introduction
%
%=========================
\section{Introduction}
Computing education is undergoing an upheaval due to Generative AI (GenAI) \cite{denny2024computingCACM} because GenAI tools like ChatGPT and GitHub Copilot can solve an impressive array of programming-related activities \cite{prather2023robots}. Some educators are incorporating GenAI into courses from the very start, giving students access to such tools, and/or using AI-first textbooks \cite{porter2023learn}. These tools could allow students to go further and faster than traditional CS1 courses have historically allowed \cite{vadaparty2024cs1llm} or have access to personalized tutoring resources at all times \cite{liu2024teaching}.

Despite the impressive capabilities of GenAI, its impact on student learning remains largely unknown. This is particularly true at the novice level because some tools, such as Copilot, were made for experts, and therefore what we know about its usefulness pertains mostly to professional developers \cite{ziegler2024measuring}. The recent ACM/IEEE-CS/AAAI CS2023 curriculum\footnote{\url{csed.acm.org}} took an optimistic outlook on GenAI, suggesting that it may be able to scaffold novice learning \cite{becker2023generative}. Indeed, early work showed that it could accelerate programming \cite{barke2022grounded}. However, since GenAI entered the mainstream, researchers have been concerned about the possibility of student over-reliance \cite{becker2023programming}. A recent blog post by Amy Ko went further, arguing that GenAI is supplanting thought and short-circuiting students' ability to reason \cite{ko2024blogpost}. 

We take this caution seriously and note that even before the advent of GenAI, researchers found that students struggled to move through the programming problem solving process due to a lack of metacognitive awareness \cite{loksa2016programming, loksa2016role, prather2018metacognitive, prather2019first}. Metacognition is a crucial skill in learning programming, which prior work has found most novice programming students lack \cite{prather2020what}. If GenAI is replacing critical thinking in programming problem solving, rather than supporting it, then it stands to reason that metacognitive difficulties faced by novice programmers could get worse with GenAI tools as well. It may also be the kind of phenomenon that is not readily apparent to instructors due to students possibly having a perception of learning while courses evaluate programming skills in traditional ways, such as through the popular ``many small programs'' approach \cite{allen2019analysis, allen2019many}.  
Therefore, evaluating novice programmer metacognition is one place to start determining whether and how GenAI is harmful to novice programming problem solving \cite{tankelevitch2024metacognitive}.

Prather et al. evaluated novice programmers in a lab study of students (n=31) solving an appropriately difficult programming problem \cite{prather2018metacognitive}. They identified five metacognitive difficulties that students faced when solving the problem and submitting it to an automated assessment tool for feedback. These metacognitive difficulties center around understanding of the problem, moving too quickly through the process, and an unwillingness to rethink the solution once the code is seemingly complete. In this paper, we replicate that study with the addition of GenAI tools to discover if the introduction of AI has solved any of these previously identified issues and what new issues it may have created. We found that all previous metacognitive difficulties remain and can be compounded by GenAI and that new metacognitive difficulties have emerged. Students with better grades and higher self-efficacy were more likely to use GenAI tools to accelerate towards a solution, while others struggled through the process, maintaining what we observed to be an unwarranted illusion of competence. We present here evidence that the GenAI tools could be widening the digital divide.

Our experiment was guided by the following research questions:

\begin{enumerate}
    \item[RQ1:] What benefits do novice programmers receive from using GenAI tools to solve programming problems?
    \item[RQ2:] What difficulties do novice programmers face while using GenAI tools to solve programming problems?
    
    %\item[RQ2] How has generative AI impacted previously identified metacognitive difficulties?
    %\item[RQ3] What new metacognitive difficulties have emerged from the introduction of generative AI programming tools?

    %\item[RQ] How have GenAI tools impacted novice programmer problem solving?

\end{enumerate}

%=========================
%
%   Related Work
%
%=========================
\section{Related Work}

\subsection{Large Language Models in Computing Education}
%\todo[inline]{Juho}

Large language models and generative AI have seen a lot of attention in computing education over the past two years~\cite{prather2023robots}. LLMs can solve most introductory programming exercises in both CS1~\cite{finnieansley2022robots} and CS2 courses~\cite{finnieansley2023my}. They can correctly answer multiple-choice questions related to introductory programming~\cite{savelka2023thrilled} and solve Parsons problems based only on an image of the problem~\cite{hou2024more}. This has raised concerns from educators about potential student over-reliance~\cite{lau2023ban,zastudil2023generative, becker2023programming,sheard2024instructor}.

Another stream of work has explored how computing instructors could utilize large language models, such as for creating programming exercises~\cite{sarsa2022automatic}, code explanations~\cite{sarsa2022automatic,macneil2023experiences,leinonen2023comparing}, and enhancing programming error messages~\cite{leinonen2023using,santos2023always,wang2024large}. Due to good performance on these tasks, there have been calls to integrate LLMs into the classroom~\cite{zastudil2023generative,cambaz2024use} (albeit with guardrails~\cite{hellas2023exploring}). Many computing education tools that utilize generative AI and LLMs have recently emerged such as \textit{CodeHelp}~\cite{liffiton2024codehelp}, \textit{CodeAid}~\cite{kazemitabaar2024codeaid}, and \textit{Promptly}~\cite{denny2024prompt}. The first two provide hints to students, while the last teaches students prompt engineering while still learning coding concepts.

In addition to exploring the capabilities of LLMs, work has started to emerge looking at how LLMs are used for code generation. Recent work has found that GitHub Copilot can increase productivity for professional programmers~\cite{peng2023impact}, although other work has found it not affecting time-on-task (even though participants preferred it over regular autocomplete)~\cite{vaithilingam2022expectation}. For professionals, Copilot use can be categorized into ``acceleration'' where they utilize it for completing a task faster and ``exploration'' where they explore potential approaches to solve the task~\cite{barke2022grounded}. In addition to these two modes of operation, novices sometimes engage in ``shepherding'' -- over-reliance where they rarely write any code of their own -- and ``drifting'' -- where they drift between Copilot's suggestions, making no progress towards the task \cite{prather2024tochi}. Regardless, preliminary results from multiple studies suggest that students enjoy having generative AI available and that the majority of them find generative AI tools helpful and do not think they over-rely on them~\cite{vadaparty2024cs1llm,liu2024teaching,pereira2024chatgpt}. The aforementioned preliminary studies are based on surveys, and thus report what students \emph{think} about generative AI and their use of it, but not necessarily \emph{how} they actually use it. Our research addresses this gap by studying how novices use generative AI tools using a rich dataset encompassing both eye-tracking and think aloud data.

\subsection{Metacognition in Programming}
Programming is about more than just writing lines of code; it involves complex cognitive processes to decompose the problem, implement a solution, and debug when necessary. As initially defined by Flavell, metacognition is the awareness and regulation of cognitive processes~\cite{flavell_1976_metacognitive}. Metacognitive awareness, along with self-regulation, are two essential components of cognitive control~\cite{schommer1990effects}, which involves students recognizing and implementing strategies to guide their problem solving and learning. This has long been known as an important part of learning programming \cite{prather2020what, loksa2022metacognition}. In a study of metacognitive awareness in novice programmers, Loksa et al.~\cite{loksa2016programming} identified six stages in the program problem solving process, which include (1) \textit{reinterpret the prompt}, (2) \textit{search for analogous problems}, (3) \textit{search for solutions}, (4) \textit{evaluate a potential solution}, (5) \textit{implement a solution}, (6) \textit{evaluate implemented solution.} They found that helping students become aware of where they were in these six stages increased performance. Prather et al. \cite{prather2018metacognitive} observed students moving through these stages and reported the metacognitive difficulties they faced (see Table \ref{tab.metacognitives}).

%Often, when students are working their way through these problem solving stages, they may experience metacognitive difficulties. Based on a study of 31 students in a classroom setting, Prather et al.~\cite{prather2018metacognitive} uncovered five metacognitive difficulties that students experienced while moving through these six stages, which include (1) \mfor, (2) \mdis, (3) \mass, (4) \mloc, and (5) \mach. \mfor \space is a difficulty faced when students form the wrong conceptual model about the right problem. \mdis \space is when an incorrect conceptual model of the problem is not corrected by re-reading the prompt. \mass \space is a difficulty related to forming the correct conceptual model for the wrong problem. \mloc \space refers to  moving too quickly between problem solving stages and experiencing a false sense of accomplishment. \mach \space is the unwillingness to abandon a wrong solution. These metacognitive difficulties were identified when observing individual students engage in a 35-minute-long programming task.

We chose metacognition as the theoretical lens through which to understand novice usage of GenAI because it allows us to measure and evaluate the usability challenges of GenAI \cite{tankelevitch2024metacognitive}. However, there is still very little known about how students actually use GenAI tools when learning programming and how that impacts their metacognitive awareness. In a recent study where students could use GenAI while learning programming, Margulieux et al. reported that students may use GenAI to support, not replace, their own problem-solving \cite{margulieux2024self}. However, they also reported that student use of GenAI tools was correlated with lower grades, lower self-efficacy, and a higher fear of failure. They suggest that this could mean LLMs are helping those who need it the most, such as students who are less prepared or not as confident in their programming abilities. They also note that it could be having the opposite effect and that lower performing students might need help using GenAI tools in ways that do not circumvent their own learning. It is possible that many lower-performing students are not aware of how GenAI is circumventing their learning.

\section{Methods}

\subsection{Context}
\begin{figure}[htbp]
    \centering
    \includegraphics[width=.7\textwidth]{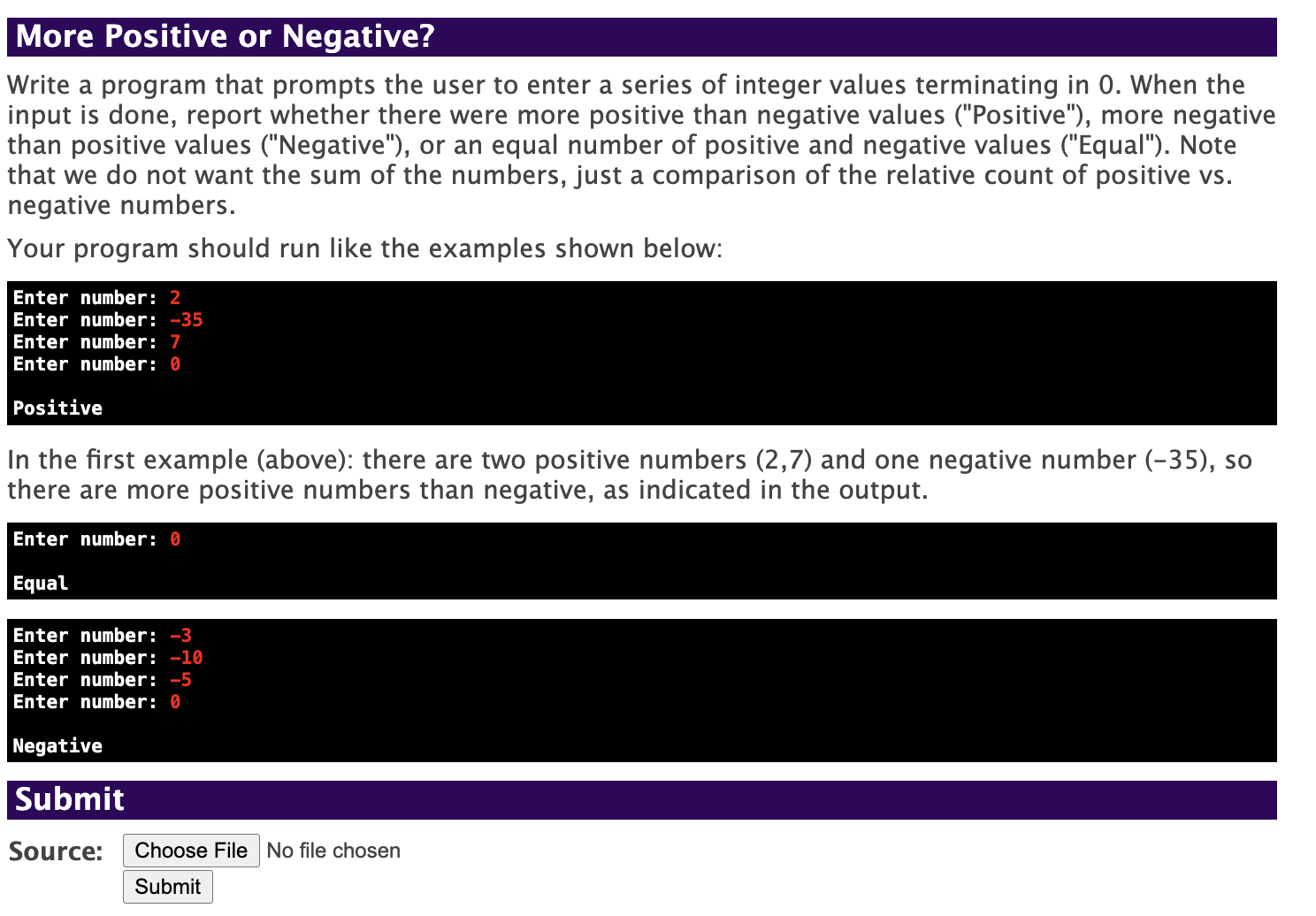}
    \caption{Problem description from Athene}
    \label{fig:atheneprob}
\end{figure}

\begin{figure}[htbp]
    \centering
    \includegraphics [width=0.7\textwidth]{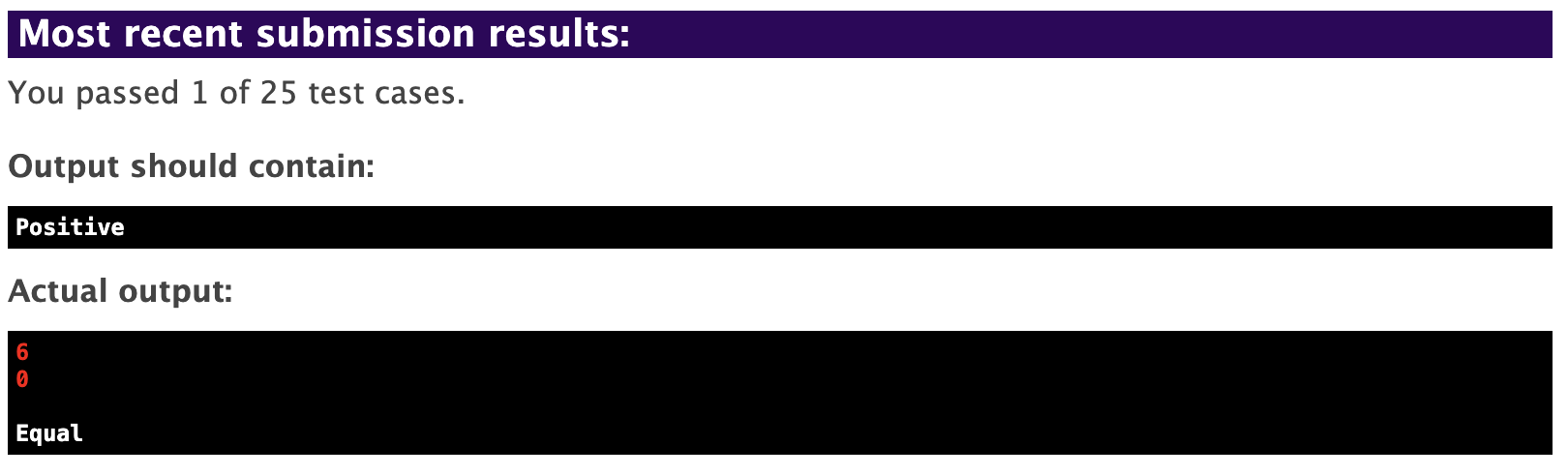}
    \caption{Sample Feedback from Athene}
    \label{fig:athenefeed}
\end{figure}

Because we were seeking to replicate the work of Prather et al. \cite{prather2018metacognitive}, we sought permission to use their automated assessment tool (AAT), Athene. This allowed us access to the same programming problem used in their previous study. This tool was then integrated into our Learning Management System, Canvas (see Figure ~\ref{fig:atheneprob}). After submission, the AAT evaluates the program by compiling it and then running it against a series of test cases. It then provides feedback to the user (see Figure~\ref{fig:athenefeed}).

The problem selected by Prather et al. \cite{prather2018metacognitive} was ``More Positive or Negative'' that determines if there are more positive or more negative numbers after an indeterminate number of integers input by the user.  Our course mirrors their course in programming language (C++) and order of topics. Loops were introduced two weeks before conducting the study (i.e. it was still somewhat new, but not so new that it was overwhelmingly challenging). We attempted to make the study setting as similar as possible to the original to increase the validity of the replication.

Unlike Prather et al. \cite{prather2018metacognitive}, whose experiment was conducted before the advent of GenAI, the course in which this study took place utilized GenAI from the beginning. The professor (one of the researchers) encouraged students to install VSCode with GitHub Copilot on the first day of class and coded live in front of students in that environment. The professor also utilized ChatGPT during class to discuss productive ways to use that tool, such as for understanding programming error messages. Furthermore, the professor frequently used class time to discuss when Copilot or ChatGPT provided unhelpful, misleading, or simply incorrect responses. The professor did this hoping that it would model positive usage of these GenAI tools for the students.

\subsection{Participants}
As in the previous study, this replication also takes place at a small research university in the USA where small class sizes are the norm. This study was approved by the lead institution's IRB as ``exempt'' research and students were provided with consent forms. There were 27 students enrolled in this CS1 course, and although extra credit points were offered for participation, only 21 students chose to opt-in. The ones who did not opt-in were offered an alternative way to receive extra credit, which was to complete an extra program of similar complexity, and one student chose to complete that requirement. As noted in Oleson et al. \cite{oleson2022demographics}, it is important to note who is present as much as who was not present and that studies in these contexts can often involve power dynamics that can alienate certain marginalized groups. Along these lines, it should be noted that one student who identified as African-American and two students who identified as Hispanic were among those that chose not to participate. From the students who participated, three students identified as African-American, two students identified as Hispanic, and one student identified as both racially and ethnically Jewish. The other fifteen identified as white or Caucasian. Additionally, two students who identified as White were from Europe, one from Germany and the other from Italy. We collected racial and ethnic identities because the institution where this study was conducted has been steadily growing towards being identified as an Hispanic-Serving Institution (HSI) and we were concerned about what issues traditionally marginalized groups may be facing. Seven of the participants identified as women and the other 14 identified as men. We collected gender identity information because previous research has shown that women are more likely to be marginalized in computing classrooms \cite{elias2022obstacles}.

% \subsection{Eye Tracking}
% To complement the think aloud data (see Section~\ref{sec:lab-study}), we also collected eye tracking data. For this study, we used Tobii eye tracking hardware and software. The hardware is a thin bar that sits at the bottom of the screen and collects gaze data from each participant. This data is then overlaid onto the screen recording of their session. It tracks both fixations and saccades and allows them to be replayed. Fixations are depicted as red circles, which grow larger as the duration of the fixation increases. The recordings of participant eye tracking were stored in an encrypted folder and only the researchers had access to it.

\subsection{Lab Study}
\label{sec:lab-study}

%Students chose a time to participate in the lab session. 
Students used a lab computer that had VSCode installed with GitHub Copilot. During the study, they were allowed to utilize GitHub Copilot as well as ChatGPT. The web browser on the lab computer had only two tabs open: the programming problem in Canvas and ChatGPT. Between sessions, the ChatGPT conversation was cleared to prevent it from using previous prompts, and the .cpp file in VSCode was deleted and remade in an attempt to prevent Copilot from leaning on previous context. 
%Unlike Prather et al. \cite{prather2018metacognitive}, we allowed local compilation before submission to the AAT. This was disallowed in the original study so as to encounter more enhanced error messages through the AAT. As this was not a focus of the present study, we decided local compilation would be appropriate. 

We followed the protocol laid out by Prather et al. \cite{prather2018metacognitive}, including pre and post checklists and guidelines and scripts found in \cite{rubin2008handbook}. First, the researcher in the room read from the script to set expectations and goals. After this, eye tracking tools were calibrated to the participant. 
We used Tobii eye tracking hardware and software. The hardware is a thin bar that sits at the bottom of the screen and collects gaze data from each participant. This data is then overlaid onto the screen recording of their session. It tracks both fixations and saccades and allows them to be replayed. Fixations are depicted as red circles, which grow larger as the duration of the fixation increases. 
Eye tracking has been utilized in programming research for decades ~\cite{behler,angelo2, angelo3} and is mostly used to measure code comprehension and debugging \cite{obaidellah}, novice gaze patterns  \cite{busjahn} and automatic skill level detection \cite{almadi}.

Next, the participant was given a warm-up task because following a think-aloud protocol can be difficult to do while also doing cognitively demanding tasks, as suggested by \cite{whalley2014qualitative, teague2013qualitative}. The warm-up task was simply writing a ``Hello World'' program while verbalizing their thoughts and explaining their actions, which should be manageable by most students at that point in the semester. After completing the warm-up activity, students were provided the ``More Positive or Negative'' programming problem (see Figure \ref{fig:atheneprob}) and asked to complete it in 35 minutes. This same time limit was used by Prather et al. \cite{prather2018metacognitive} because a majority of students could solve it within that time frame when used as an in-class quiz. While the participant worked on the programming problem, researchers took notes on everything the participant said and did, while minimizing interactions with the participant as suggested by best practices \cite{ericsson1993protcol}.

Afterwards, we asked a series of questions about their perceptions on how helpful Copilot and ChatGPT were during the problem solving, their perceptions about AI in general, how much they use AI, their prior experience using AI and prior experience programming, family socio-economic status, self-described race and gender, how much they work, age, and major. One week later, we also collected self-efficacy data from students at the beginning of class using the self-efficacy subscale of the MSLQ \cite{pintrich1991manual}. We did this so that we could get their self-efficacy perceptions apart from the lab, which could be stressful to some participants.

\subsection{Data Analysis}
Participants were given unique random identification numbers. Participant weekly programming quiz grades were deidentified and added to the data according to their identification number. Participant observation notes were then compiled into a spreadsheet, split by line, for tagging. Researchers compiled the following initial codebook of 15 codes: six stages of programming problem solving by Loksa et al. \cite{loksa2016programming}, five metacognitive difficulties by Prather et al. \cite{prather2018metacognitive}, verbalized positive self-efficacy, verbalized negative self-efficacy, verbalized positive emotion, and verbalized negative emotion. 

Two researchers then met and discussed tagging procedure and initial ideas of what each of the tags meant. The two researchers then separately coded the first participant observation session, tagging every line with at least one of the programming problem solving stages. As they tagged each line, they also watched the video playback of the session with eye tracking data overlaid (see figures below for examples). This enhanced the researchers' abilities to know what a participant was looking at when they said or did particular things, how long they spent looking, and what they chose to look at that may not have been verbalized. The two researchers then met together to discuss their disagreements and resolved them through discussion. This process was repeated four times until, after the fourth participant data were tagged this way, the researchers reached a Cohen's Kappa of 0.74, which is considered high agreement \cite{mchugh2012interrater}. After reaching this level of agreement from tagging separately, the researchers tagged the rest of the data independently. 

As the researchers tagged all data, they discussed any occurrence where they thought new phenomena had arisen out of the raw data. In this way, three new metacognitive difficulties were identified and therefore were added as tags during tagging: \mint, \mmis, and \mpro.

\subsection{Limitations}
One limitation of our study is the relatively small sample size, with only 21 participants compared to 31 participants in the original study by Prather et al.~\cite{prather2018metacognitive}. While this smaller sample size may limit the generalizability of our findings, it is important to note that our study incorporated additional analyses with eye tracking technology, which provided novel insights into the metacognitive difficulties experienced by participants in the study.
There were also limitations associated with the generative AI tools that were used in our study. While recent survey studies have shown that ChatGPT and GitHub Copilot are both being used more frequently by students~\cite{hou2024effects, prather2023robots}, it is important to acknowledge that these are not an exhaustive representation of the breadth of generative AI tools that are available. Despite our efforts to compensate for the smaller sample size through robust statistical analysis and in-depth contextualized qualitative analysis, it is imperative for future research to conduct studies with larger sample sizes encompassing diverse generative AI tools to corroborate and extend our findings. Finally, we were limited by the single site of the experiment as noted by Oleson et al. \cite{oleson2022demographics}.

\begin{figure*}[htbp]
\centering
 \includegraphics[width=1\linewidth]{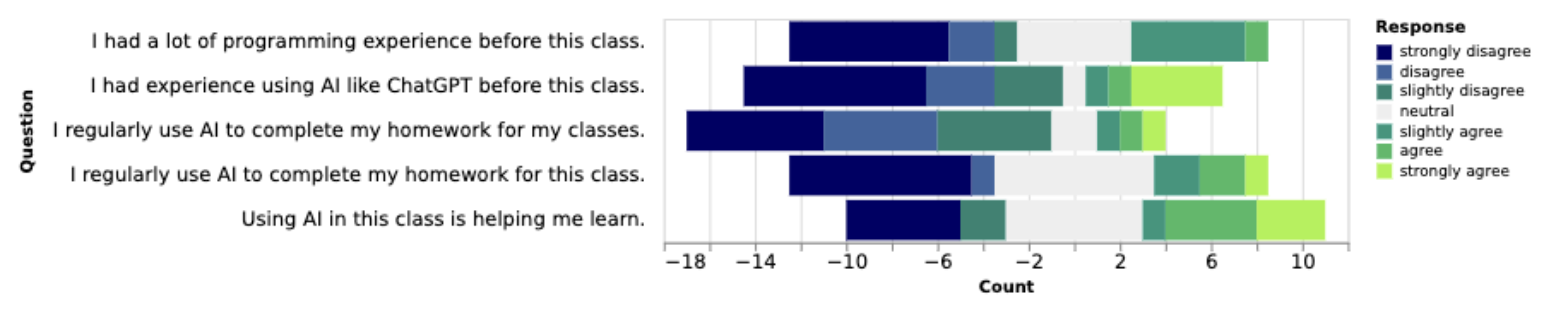}
 \caption{Interview Questions about programming and AI experience}
 \label{fig:demographics1}
\end{figure*}

\section{Results}
Twenty-one students participated in the lab session. All students except for one completed a working program within the time limit.  Times varied from 5 to 35 minutes and averaged 17.1 minutes with a standard deviation of 8.1. Students also accepted suggestions from Copilot at rates ranging from 10\% to 53\%. See Table \ref{tab.bystudent} and Table \ref{tab.metasnometas} for summary data on these measures. Likert interview question results are shown in Figure \ref{fig:demographics1}.

% Count of times we tagged each stage, Stage state changes? maybe next time
% tab DATA rows 780-805, 'efficacy' produces the csv

Nine students were tagged with metacognitive difficulties taken from previous work \cite{prather2018metacognitive}, while eight students were tagged with new metacognitive labels (see Table \ref{tab.metacognitives} for definitions). Only 1 student was tagged with new metacognitive label without having any old labels. We provide the full data of which student encountered each metacognitive difficulty in Table \ref{tab.bystudent}.

%Pearson correlations were high (.7 - .89) for two variables and moderate (.5 - .69) for a eight variables.

Several variables were correlated with \M-d, though we do not discuss all for space constraints (see Table \ref{tab.pearson_stats}). Student grade and new \m-d were moderately\footnote{r 0.0-0.19 very weak, 0.2-0.39 weak, 0.4-0.59 moderate, 0.6-0.79 strong and 0.8-1.0 very strong.} negatively correlated (r= -.503, p= 0.020).  It is reasonable to imagine that students with lower grades would have more difficulties than students with higher grades.  Although with less significance, old \m-d were similarly correlated with grade (r= -0.48, p= 0.268). \M-d counts and negative self efficacy were moderately correlated (r= 0.552, p= 0.009). It is not surprising to consider that a student who had low self-efficacy would also experience more metacognitive issues than students with high self-efficacy. Higher self-efficacy scores were associated with higher course grades (r= 0.552, p= 0.009), as also seen in other studies on the role that self-efficacy plays in computer science course performance \cite{Ramalingam2004}. The longer a student took to solve the problem, the more \m-d they were likely to exhibit (r= 0.693, p= 0.00049).  This was not surprising, given that the \m-d interfere with successful problem solving.  Relatedly, the longer a student took to solve the problem, the lower their course grade tended to be (r= -0.727, p= 0.0002).

Programming experience was strongly correlated with AI experience (r= 0.622, p= 0.0026), though these were both self-reported.

\begin{table*}[htbp]
\centering
\caption{Pearson Correlations for moderately and highly correlated variables}  
\label{tab.pearson_stats}

\begin{tabular}{llrr}
\toprule
  Variable & Variable & r & p \\
\hline

%Metacognitive issues count & Old metacognitive issues count & 0.818 & 0.00000 \\
%Metacognitive issues count & New metacognitive issues count & 0.818 & 0.00000 \\
%Old metacognitive issues count & New metacognitive issues count & 0.806 & 0.00000 \\
New metacognitive issues count & Time & 0.729 & 0.00020 \\
Course grade & Time & -0.727 & 0.00020 \\
Metacognitive issues count & Time & 0.693 & 0.00050 \\
Used AI in class & Finds AI helpful in class & 0.661 & 0.00110 \\
Programming experience before class & Experience with AI & 0.622 & 0.00260 \\
Old metacognitive issues count & Time & 0.621 & 0.00270 \\
Course grade & Self-efficacy score & 0.553 & 0.00940 \\
Problem states & Time & 0.552 & 0.00940 \\
Metacognitive issues count & Negative self efficacy count & 0.552 & 0.00940 \\
Course grade & New metacognitive issues count & -0.503 & 0.02010 \\
Course grade & Old metacognitive issues count & -0.482 &0.02682 \\
\bottomrule
\end{tabular}
\end{table*}

\subsection{Participants Who Struggled}
Although only one student failed to complete the task, half of the students were observed struggling for various reasons, including exhibiting one or more \m-d. We found each of the metacognitive difficulties previously identified by Prather et al. \cite{prather2018metacognitive}: \mfor, \mass, \mdis, \mloc, and \mach. But in addition to those, we also found three new metacognitive difficulties centered around use of GenAI: \mint, \mmis, and \mpro. See Table \ref{tab.metacognitives} for a full description of each metacognitive difficulty both from previous work and our results.

We describe the experiences of these participants below using observation notes, student verbalizations, and the recorded replay and gaze interactions with VSCode, Copilot, and ChatGPT, informed by the presentation of similar data in \cite{angelo3,busjahn}.

\begin{table*}[htbp]
\centering
\caption{Definitions of Old and New Metacognitive Difficulties}  
\label{tab.metacognitives}

\begin{tabular}{p{2cm}p{9cm}}
\toprule
Name & Description \\
\hline
 \\
\multicolumn{2}{c}{\textbf{Previous} \cite{prather2018metacognitive}} \\
 \\
\hline
\mfor	& Forming the wrong conceptual model about the right problem. \\
\mdis & Dislodging an incorrect conceptual model of the problem may not be solved. \\ 
\mass & Forming the correct conceptual model for the wrong problem.  \\ 
\textbf{Location} & Moving too quickly through one or more stages incorrectly leads to a false sense of accomplishment and poor conception of location in the problem-solving process. \\ 
\mach & Unwillingness to abandon a wrong solution due to a false sense of being nearly done. \\ 
\hline
\\
\multicolumn{2}{c}{\textbf{New} } \\
\\
\hline

\mpro & Being conceptually behind in the course material but unaware of it due to a false sense of confidence \\
%Students were unprepared for the learning activity because they lack fundamentals. They are conceptually behind the current location in the course and do not understand that they have not progressed to the point where they should be.  \\

\mint & An inability to concentrate on problem solving due to frequent interruptions and code suggestions. \\
%A repeating pattern where a GenAI tool interrupts a student with suggestions that take them out of the flow of problem solving.  \\ 

\mmis & The tool leads the user down the wrong path.  \\
\bottomrule
\end{tabular}
\end{table*}

\begin{table*}[htbp]
\centering
\caption{Count of types of metacognitive difficulties, Completion time, and Accept-Rate}  
\label{tab.bystudent}
\begin{tabular}{clcrc}
\toprule
anonid & Difficulties & Count & Time & Accept Rate \\

1 & Forming, Dislodging, Location, Achievement, Progression & 5 & 35.00 & 0.18 \\
2 &  & 0 & 6.80 & 0.10 \\
3 &  & 0 & 12.80 & 0.31 \\
4 & Dislodging, Assumption, Location, Achievement, Progression & 5 & 22.83 & 0.22 \\
5 &  & 0 & 10.20 & 0.28 \\
6 &  & 0 & 7.60 & 0.22 \\
7 & Location, Interruption & 2 & 28.17 & 0.17 \\
8 & Location, Interruption & 2 & 19.75 & 0.20 \\
9 & Location & 1 & 11.90 & 0.31 \\
10 &  & 0 & 19.13 & 0.24 \\
11 & Mislead & 1 & 20.93 & 0.37 \\
12 &  & 0 & 12.43 & 0.31 \\
13 & Forming, Progression & 2 & 18.35 & 0.53 \\
14 & Forming, Achievement, Progression & 3 & 16.23 & 0.50 \\
15 &  & 0 & 11.76 & 0.17 \\
16 &  & 0 & 26.48 & 0.23 \\
17 & Location, Interruption & 2 & 29.11 & 0.31 \\
18 &  & 0 & 10.50 & 0.28 \\
19 &  & 0 & 5.00 & 0.33 \\
20 &  & 0 & 9.70 & 0.29 \\
21 & Forming, Mislead & 2 & 24.90 & 0.46 \\

\bottomrule
\end{tabular}
\end{table*}

\begin{table*}[htbp]
\centering
\caption{Acceptance Rates of Copilot Suggestions by Metacognitive issues present}  
\label{tab.metasnometas}
\begin{tabular}{lcc}
 & Metas & No Metas \\
 \toprule
mean & 34.1\% & 24.5\% \\
stddev & 12.5\% & 6.6\% \\
min & 17.0\% & 10.0\% \\
max & 53.0\% & 33.0\% \\
\bottomrule
\end{tabular}
\end{table*}

As seen in Table \ref{tab.metasnometas}, students who experienced \m-d tended to accept CoPilot suggestions at higher rates.  Students who are less prepared have a more difficult time determining whether a CoPilot suggestion is worth accepting.  They accepted more suggestions that they subsequently reworked or rolled back entirely.

\subsubsection{P1: \mfor, \mdis, \mach, \mint}

Although the problem description explicitly states not to sum the numbers, P1 first wrote code that would actually sum them. But from their verbal statements, it appears they intended to solve the correct problem and were simply incorrectly implementing it, indicating a difficulty described in the previous work as \mfor. As they struggled to write this summing solution, Copilot would regularly interrupt, causing them to pause and consider each of its suggestions. In response to these interruptions, they said things like ``These prompts are distracting sometimes'' and ``I’m trying to think of...never mind, wait'' and ``No, go away, please stop what you are doing.'' They often bounced between implementing a solution and evaluating Copilot's suggestions, which led to several \mint \space difficulties in the process. 
% but we claimed above that his was "Forming".  How to connect and explain these two?
The disconnect between their working conceptual model of the problem and a successful approach was not immediately apparent to them even though they went back to the problem description to re-read it several times.  This interaction of being sent back to the problem description while being "in the weeds" we call \mdis. P1 rewrote the solution several times, first with conditionals and then with nested loops, leading to observed and verbalized frustration.

%\begin{figure*}[htbp]
%\centering
%  \includegraphics[width=1\linewidth]{Figures/eyetracking/P1-20min-first-chatGPT.png}
%  \caption{P1's first use of ChatGPT, looking at keyboard at the time, i.e. no on-screen gaze.}
%  \label{fig:P1}
%\end{figure*}

After trying and failing to get their code to compile, they sought out ChatGPT, saying, ``Okay, I’m using ChatGPT now. I love ChatGPT. It's so helpful.'' They copied and pasted all of their code into ChatGPT without giving it a prompt, so the tool replied with a simplified version of their incorrect implementation, to which they replied, ``Oh, so I don't need to put that. They made this so simple, it’s annoying. Oh wait, no, that’s not what I asked for. I need to figure out what to prompt it. Or I can just copy and paste the Athene.'' Realizing they needed to provide more context for ChatGPT, they instead pasted the problem description into ChatGPT as a subsequent prompt in the same conversation thread. Without reading the suggestions provided by ChatGPT, they immediately began comparing its code to their own code. Further Copilot interruptions continued and they said, 
``Oop please go away.'' ChatGPT suggested a do-while loop, which had not been covered in class, further compounding comprehension issues. Looking at ChatGPT's suggested code, they added an additional for-loop above the nested for-loop and then started a new ChatGPT thread asking ''why isn't my code outputting the negative or positive?'' Although the suggestions from ChatGPT were wildly different from their own code, they continued trying to use their solution and fix it. Although major structural changes were necessary to the code, they were unwilling to abandon their solution, showing an \mach \space difficulty and saying, ``Maybe if I put the numbers differently...nope it’s just difficult.'' 

Copilot continued to interrupt and finally they decided to submit to Athene to see if that would provide some new help. After that provided no new insight, they copied the problem description into ChatGPT, from which they deduced that they needed two new variables at the top to track positives and negatives. This was all the information that Copilot needed to start providing extremely useful suggestions, which they began to accept. They then deleted nested loops and started patterning the code after what was provided by ChatGPT. Although close, they were unable to arrive at a working solution before the time ended, saying, ``I love how I spent 30 minutes on different code and then got it in a few minutes with this.'' 

When asked in the post-interview whether they thought Copilot was helpful, they said, ``For the most part, yes. It was great when it gave me the right answer but when it didn't it was distracting me and throwing me off.'' Similarly, when asked if ChatGPT was helpful, they said, ``Yes. I put my code in and it didn't give me the right answer at first but that was my fault. Then I put the prompt in and scanned it and put some pieces into my code.'' Their responses reinforce that there seems to be a disconnect between their ability to solve the problem and how much help GenAI tools actually provided.

%P4: Assumption, Dislodging, Location, Achievement, Interruption
\subsubsection{P4: \mass, \mdis, \mloc, \mach, \mint}
P4 began by carefully reading through the problem description, twice. They then immediately started coding by writing a for-loop that would iterate ten times. Unhappy with this, they deleted it and started over with a while loop that would iterate until the input was zero. Inside the loop, they started setting up a conditional. As Copilot gave constant suggestions, they said things like, ``Stop! That’s so distracting'' and ``Oh, shut up.'' This occurred throughout the session, showing an \mint \space difficulty. Although Copilot suggested generic and largely unhelpful conditions, they typed a condition to use the modulus to determine whether the number is even or odd, showing an \mass \space difficulty. They then added variables at the top for counting evens and odds. With this information, Copilot began suggesting long blocks of code to solve the even or odd problem, which they accepted, giving them nearly an entire solution for that (incorrect) problem. Despite multiple looks back at the problem description, they continued formatting the code for the even or odd problem instead. After testing it and receiving incorrect responses, they read the code and the prompt multiples times, confused, saying ``Okay wait that doesn’t work. Why?'' Stuck at the end of the implementation phase of the problem solving process, they did not realize that they had skipped most of the early stages of the programming problem solving process and jumped right to coding a solution, showing a \mloc \space difficulty.

%\begin{figure*}[htbp]
%\centering
%  \includegraphics[width=1\linewidth]{Figures/eyetracking/P4-11min-abs-function.png}
%  \caption{P4 trying to use an absolute value function after asking ChatGPT.}
%  \label{fig:P4}
%\end{figure*}

They turned to ChatGPT for help and prompted, ``How to find absolute value in C++?'', apparently deciding that they needed this to fix their code. After reading the code response from ChatGPT, they wrapped their conditional checking if a number is even with a call to the built-in absolute value function. More testing revealed the same incorrect responses and they re-read the problem prompt again for the fifth time, showing clearly a \mdis \space difficulty. Frustrated, they returned to ChatGPT and prompted, ``Why wouldn't this code work correctly given the following values?'' and pasted in their code and the test cases displayed in the problem description. ChatGPT replied that they had not initialized the even or odd variables and that this could lead to undefined behavior in C++, to which the student replied excitedly, "Oh! This isn't C\#! I can’t believe I did that.'' They initialized one variable, recompiled, and tested again, saying, ``Okay, it works now.'' With code that looked nearly done and Athene providing feedback that helped them feel as if they were getting closer to completion, they did not think of changing their solution at the fundamental level that it needed, showing an \mach \space difficulty.

Submitting to Athene again, they still received incorrect output. Believing it to be a trivial spacing issue, they made those changes, tried again, and received the same result and asked, ``Why is this not working?'' Finally, they re-read the problem description again and exclaimed, ``OH. I’ve been doing this wrong the entire time. Positive or negative numbers! Oh my gosh!'' They immediately went back to the code and started editing it. Copilot attempted to suggest more code related to the even or odd problem, to which they replied, ''Stop. Okay, I can't believe this.'' As they worked on editing to align with the correct problem, they started typing the even or odd conditions again, but realized it and exclaimed, ''OMG! Why did I do that again!'' As they worked on the correct solution, Copilot assisted toward the goal and they said, ''Why did I not catch that until the end?'' Within three minutes of realizing they had been solving the wrong problem, they arrived at a correct solution.

When asked in the post-interview whether they thought Copilot was helpful, they replied, ``No, it kept getting in my way when I was trying to think. It was interrupting my thought process.'' But when asked about ChatGPT, they replied, `It can provide me help as I'm learning code. Chat does some critical thinking about what I'm asking and can understand the problems I'm asking it.'' From their comments, they seemed to think that ChatGPT had \textit{augmented} their critical thinking rather than \textit{replaced} it, but the data above contradicts that.

\begin{figure*}[htbp]
\centering
  \includegraphics[width=1\linewidth]{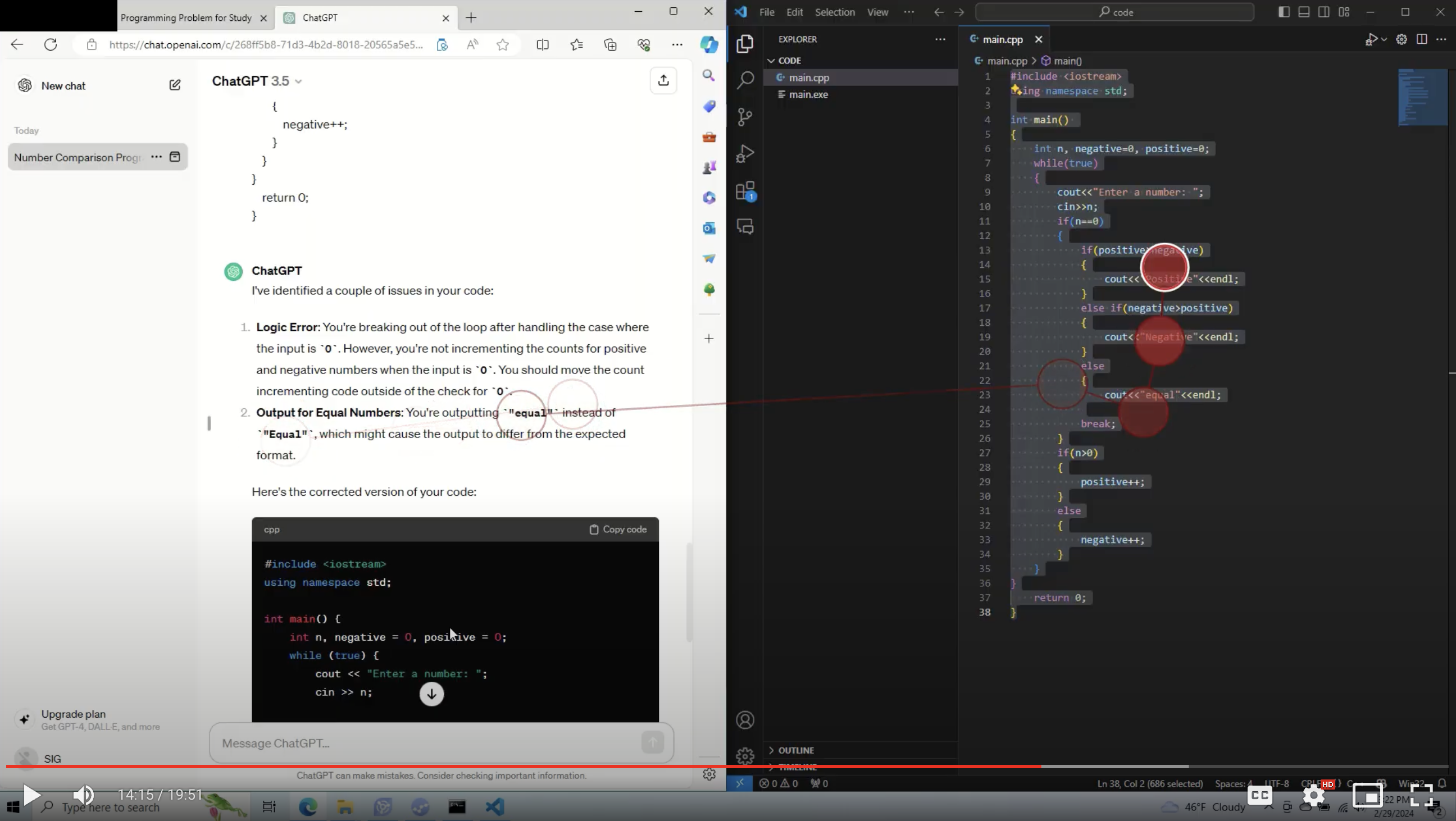}
  \caption{P8 consulting ChatGPT to find the logic error with their code.}
  \label{fig:P8}
\end{figure*}

%P7: Location and Progression.
\subsubsection{P7: \mloc \space and \mpro}
P7 started by thoroughly reading the prompt and then trying to start coding. Copilot immediately provided a long suggestion and they seemed overwhelmed, saying, ``Woah, interesting. The first thing I’ll do is pull up ChatGPT.'' They then opened ChatGPT and began writing a prompt, saying, ``I use ChatGPT more like a personal tutor than an answer solver.'' The prompt asked ChatGPT: ``help give me instructions but not the full answer on how to write code in C++ that keeps track of numbers on whether they are positive or negative at the end.'' ChatGPT provided a series of six step-by-step instructions followed by a suggested code solution. P7 started to read these instructions but stopped after the first one and scrolled down to the code. They then copied the code from ChatGPT into their own file, saying, ``So it’s giving me some code. I’m going to copy it so I can understand what it’s saying.'' 

However, the solution generated simply counted and reported the number of positive and negative numbers entered by a user, rather than which one was input most frequently. Furthermore, the solution contained extra constructs not introduced in class that confused them, such as ``std::'' before output and input statements as it did not use the standard namespace. They compiled and ran the code anyway and found that it did not output exactly as shown in the problem description, so they attempted to make the output lines match. They did not understand the generated solution, which used a while(true) loop, as evidenced by asking the researcher, ``How does a loop stop?'' Without building this solution, and instead just taking what ChatGPT generated, they had skipped crucial steps in the programming problem solving process, and were now lost, showing a \textit{Location} difficulty. They then made several changes to the code to remove unfamiliar constructs, which caused many programming error messages to appear upon compilation. Confused, they undid these changes.

%\begin{figure*}[htbp]
%\centering
%  \includegraphics[width=1\linewidth]{Figures/eyetracking/P7-16min-ChatGPT.png}
%  \caption{P7 consulting ChatGPT and comparing its code to his code.}
%  \label{fig:P7}
%\end{figure*}

Unable to determine how to proceed, P7 copied the code back into ChatGPT and prompted, ``I'm needing my code to know whether there are more positive or negative, or equal number of numbers when having put in a list of numbers for example.'' 
%(See Figure \ref{fig:P7}) 
They read the code that ChatGPT generated, looked back over their own code, and then copied it into their source file, saying, ''I see...I like this one more.'' After formatting the output strings to be exactly like the required output in the problem description, they compiled and ran the code, commenting that it was only missing an extra newline between pieces of the output. They attempted to add this, incorrectly, and once again consulted ChatGPT for a solution. Once it was provided, they read it and said, ``Oh I was just forgetting the cout. That's all I forgot, I knew that was the right place.'' 

Although converging toward a solution, P7 did not understand how to construct the solution, did not understand basic constructs used in the solution, and struggled to fix basic errors, revealing a \mpro \space difficulty. Finally, after securing the last bit of necessary code from ChatGPT, they were able to submit the program to Athene and pass all test cases.

\subsubsection{P8: \mloc \space and \mpro}
P8 began by carefully reading the problem description and then created a variable for accepting input. They next typed the ``while'' keyword and Copilot generated a loop that would sum responses. P8 read this carefully and decided that it wasn't what was needed, so they ignored it and instead used a while(true) loop. Copilot continued to provide suggestions, which the participant mostly ignored, instead constantly referring back to the problem description and slowly typing out their solution, saying, “I can use Chat[GPT] but I want to try and solve it by myself first." They added console input and a conditional statement to break the loop if that input was zero. They tried to add a for-loop, but didn't know what to put inside the parenthesis. By skipping crucial problem solving stages and jumping right to implementation, they found themselves stuck with no clue how to proceed, indicating a \textit{Location} difficulty. 

So, they copied the problem description and put it into ChatGPT without adding specific instructions. ChatGPT generated a complete solution and P8 read part of it and immediately began augmenting their code, starting first with variables to track positive and negative numbers, saying, ``I feel kinda stupid for not thinking about that'' and ``I gave up too soon. I am a failure.'' They deleted the for-loop and Copilot now suggested the code needed to increment the new counters, which they accepted. They added code to output the results inside the breaking condition in the  while loop, which Copilot also helped complete. However, the brackets were not aligned properly and Copilot suggested code without a closing curly brace. When this produced a compiler error, they resolved it by adding a curly brace at the line number provided by the programming error message, which was not the correct place to insert it. Although the  code compiled, it did not pass any test cases.

Stuck again, they asked ChatGPT ``what is wrong with my code'' without providing any code. ChatGPT responded with some basic steps to help, which included ensuring variables were initialized and input validation. Ignoring this, they copied their code into ChatGPT and asked the question again. This time, it responded with two issues: a simple formatting inconsistency and the logic error created by the misplaced curly brace (See Figure \ref{fig:P8}). P8 carefully read the text explaining both errors and immediately fixed the formatting problem, but did not understand how to fix the logic error. At this point, unable to proceed but having struggled through the use of using loops effectively aided only by Copilot and ChatGPT, P8 displayed a \mpro \space difficulty with their comprehension of course topics several weeks behind.

Finally, they began changing code to explicitly mirror that which was provided by ChatGPT, submitted, and passed the text cases. When asked if they thought Copilot was helpful during the lab, they said, ``It was absolutely helpful but it would sometimes throw me off because it would suggest other things that were not necessary. After clarifying with chat, I was right, but was initially confused because I thought that Copilot was giving me the right answer.'' They had a similarly helpful picture of ChatGPT, saying, ``Whenever I got stuck, it provided me with the option for what I needed to do. While it doesn't always fully understand the prompt, it can help you get going in the right direction. It helped me validate my initial problem solving.'' Their post-test comments, however, seem to contradict what they said and did during the lab itself where they used ChatGPT to provide a solution to the problem, not to validate their own ideas.

\subsubsection{P9: \mloc}
P9 began by carefully reading the problem description and test cases. They then began implementing their solution by creating a variable and asking for input. Copilot suggested a set of conditional statements that would output whether the number was positive, negative, or zero, which they accepted. They then declared and initialized variables at the top to count positives and negatives, replacing the output statements with incrementing the correct variable. They then moved on to create a conditional statement at the bottom of the  code. Copilot correctly suggested the set of conditions that would produce the correct answer. However, user input would only occur once as there was no loop. P9 skipped crucial problem solving planning stages, jumping directly to coding and was enticed by Copilot into quickly producing code, and therefore displayed a \textit{Location} difficulty.

As they cleaned up the code, they deleted the ``if'' in ''else if'' but left the condition, leading to a compiler error. The programming error message said they needed a semicolon before a curly brace, which of course was not helpful. Unsure how to proceed, they pasted the error (only, no code) into ChatGPT, saying, ``I’ll just see if it will tell me where I’m missing one.'' ChatGPT replied with some boiler plate code and suggested a semicolon directly after a curly brace. Unsure how to proceed, P9 pasted their code into ChatGPT and asked it where the code was missing a semicolon. ChatGPT responded with the correct answer this time, explaining the issue and fixing the code. Seeing at last that they needed a while loop, P9 edited the code to include a loop, submitted it to Athene, and passed all test cases.

When asked about their use of Copilot during the lab session, they replied, ``Yeah, it just sped it up but I could have eventually figured it out if I didn't have it. It tends to be more distracting to me so I don't use it when I'm doing it by myself.''

\subsubsection{P11: \mmis}
P11 carefully read the prompt and verbalized plans to begin structuring their code with the sample output as a guide, declared variables, and selected ``a while loop instead of for loop because the number of times is unspecified.'' When they began typing a while loop, Copilot suggested the correct condition and they accepted it. However, inside the loop it suggested a line with a variable not even declared: ``total+=num;'' They then went back to the top of the main function to declare the total variable. Copilot's suggestion here, and their subsequent acceptance of it, reveals a \mmis \space difficulty. Unaware that Copilot's suggestions are not leading them toward an algorithm that can solve the problem, they continue down this path, getting closer to a seemingly correct solution. Although they are not solving the problem correctly, it is not a \mfor \space difficulty because it wasn't their idea; it was Copilot's suggestion.

They next decided to update the solution to have variables to track both positive and negative totals and then accepted a Copilot suggestion for series of conditionals below the loop for output. The code compiled, but when run created an infinite loop. After trying to make changes and getting the same result, they sought help from ChatGPT. ChatGPT responded with code and without reading it they added, ``only a hint.'' This time ChatGPT gave only text and replied that the code was not getting more input inside their loop. After making several ineffectual and unnecessary cosmetic changes, such as indentations, they looked to the problem description again and discovered they had been solving the problem incorrectly.

\begin{figure*}[htbp]
\centering
  \includegraphics[width=1\linewidth]{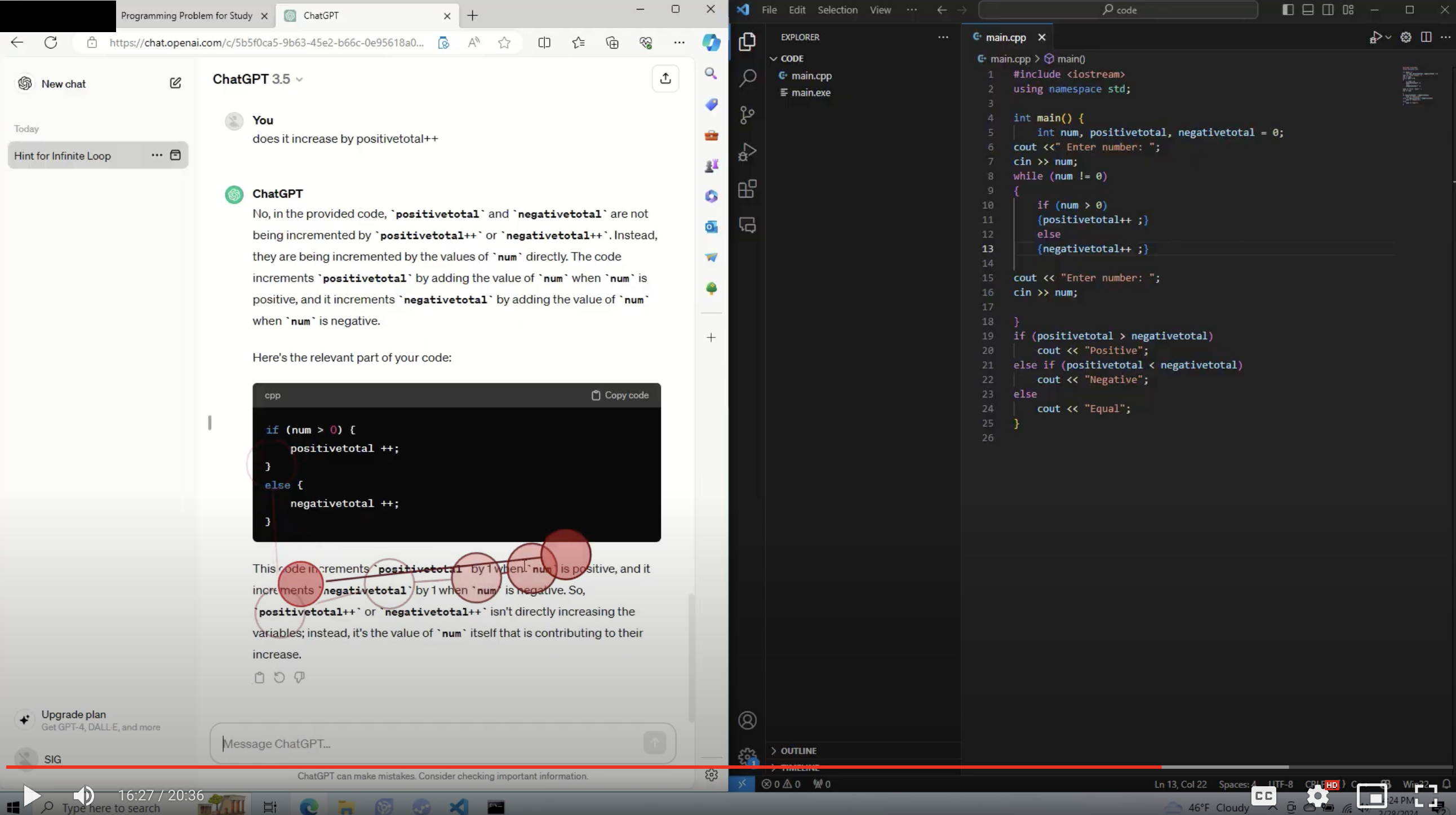}
  \caption{P11 reviews a misleading reply from ChatGPT.}
  \label{fig:P11}
\end{figure*}

After changing the arithmetic in the loop to increment, rather than adding the value to the total, they compiled and tested again. However, it still did not correctly output expected values from test cases, so they copied and pasted it into ChatGPT. ChatGPT responded with congratulations on fixing the code because it no longer had an infinite loop. They then had to specifically ask why it is not counting correctly. This time it told them to pay attention to how they were updating them inside the loop. Unbeknownst to them, the hint was incorrect and they merely needed to initialize the variables. However, they asked ChatGPT for a more detailed hint, to which it replied with a longer version of the previous unhelpful hint. They then asked ChatGPT ``does it increase by positiveTotal++?'' ChatGPT responded with an extremely misleading reply based on previous versions of their code, telling them to increment by the user input instead as they had previously done before realizing they had been solving it incorrectly.
% (see Figure \ref{Fig:P11}).

Still confused, they pasted their code into ChatGPT again. This time it responded beginning with ``Your updated code looks correct.'' They typed, ``It is still not counting negative inputs.'' Finally, ChatGPT responded helpfully letting them know that C++ can output garbage values when variables are not initialized. After initializing their variables, they compiled and tested again, and then submitted a correct solution.

When asked about their use of Copilot during the lab session, they replied, ``I think it was helpful but I think it can be frustrating because it can take away from my own thinking. But it saves me time by giving me suggestions that I can use or not.''

\subsubsection{P13: \mfor \space and \mint}
P13 began by reading the first few lines of the problem prompt and then verbalized plans to build their program, saying, ``Okay so now I understand it. I’m going to make a while loop, and while the num is not equal to zero.'' However, they quickly revealed a \mfor \space error, saying, ``I’m just trying to figure out how I’m going to do this. Okay ummm so I need to figure out how to create more loops inside the while loop, I think.'' As they worked, Copilot began suggesting many different types of solutions, including one to the even or odd problem, which they accepted and then deleted. The successive interruptions, showing an \mint \space difficulty, led them to pick several different solution strategies before they finally quit and sought the help of ChatGPT. ``I’m just going to use ChatGPT to give me a hint.''

They pasted the full problem description as well as their code into ChatGPT, prompting it for a hint. ChatGPT replied with three suggestions: variables, counting the positives and negatives inside the loop, and comparing the values at the end for output. This was a good blueprint for the program and they used it to structure their code into a nearly correct solution. However, they took input from the user before entering the loop and also again at the top of the loop, so that the first number would always be overwritten. They asked ChatGPT again for help, this time comparing the code it provided to their own, and discovered the issue. They fixed it, submitted to Athene, and completed the problem.

When asked about their use of Copilot during the lab session, they said, ``I don't think so. I think it just kinda overloaded my brain. I would be thinking about something and it would give me something else to think about and then I would forget what I was actually thinking about.''

\subsubsection{P14: \mfor, \mach, \mint}
%P14 is Forming, Achievement, Interruption. 14 has an "if-loop" (cite lit on if-loops). Her stated conceptual model is that it will ask the user for more than one number. When it doesn't, she's confused.
P14 began by reading the problem description. They then began coding, verbalizing a plan, ``Start by declaring an integer, we’ll call it num. The first thing is cout enter number...thank you Copilot. Then cin num.'' Copilot provided some assistance, which they accepted. They said, ``I guess that needs to be in a loop because we’re doing it over and over again? I guess so. I guess there has to be an if-statement somewhere.'' They then typed a conditional statement and put another conditional inside it, treating the outer conditional like a loop. Their stated conceptual model is that it will ask the user for more than one input, displaying a \mfor \space difficulty. After re-reading the problem description, they said, ``Need another if-statement if number is equal to zero. Yes, copilot is helping me, but not much.'' Copilot interrupted over and over again, causing them to stop and read these prompts each time, showing an \mint \space difficulty.

%\begin{figure*}[htbp]
%\centering
%  \includegraphics[width=1\linewidth]{Figures/eyetracking/P14-10min-if-loop.png}
%  \caption{P14 after changing the \textit{while} keyword back to an \textit{if}.}
%  \label{fig:P14}
%\end{figure*}

After adding some comments and formatting, they said, ``So, it runs it until it gets a zero.'' They then added the conditionals for output below. In doing so, they decided to delete the outer conditional statement and then test it. When it failed, they verbalized confusion, saying ``I know it could be a while or an if, I’m not sure which one. Let’s try a while loop. Ok, I understand now.'' However, testing this revealed an infinite loop, so they replaced the while keyword with if, returning to their previous ``if-loop'' design. They added an extra input statement at the bottom of the outer conditional, ran it, and were confused, saying, ``I only input twice. Why did it...even though what I said was...even though I said three.'' They spent a couple minutes tracing teh code, saying, ``The number was not zero. So why did it take me out of the if-statement?'' At this point, the solution looked nearly done, could be run and receive input, and P14 seemed blind to the fundamental error in their code, displaying an \mach \space difficulty. 

They returned to the problem description and tried several other changes  fix the issue, none of which helped. Finally, they said, ``I still don’t know what this means. Let’s ask ChatGPT.'' This helped them realize that the outer conditional needed to be a loop. They changed it and finished the problem.

\subsubsection{P17: \mloc \space and \mpro}
P17 began by carefully reading the problem description. They then made a variable to accept user input and prompted the user for that input. After this, however, they stalled for a couple of minutes, saying, ``Trying to get familiar with what it's asking.'' They decided to ask ChatGPT, saying, ``Finding a way to word it to Chat[GPT] so it can help to have a side by side on what it gives me and my work to check myself.'' They pasted the first paragraph of the problem description into ChatGPT and it replied with a program in Python. They replied to it that they needed it in C++, which it then provided, although it used a do-while loop that had not been covered in class. After spending two minutes reading the code from ChatGPT, they said, ``Trying to figure out how many variables I need to use. I need a positive, negative, and num. Trying to figure out why we need 3 variables.'' P17 skipped several early stages and was stuck in implementation without a viable plan, revealing a \mloc \space difficulty.

They tabbed back to the problem description, added new variables and then attempted to continue writing the solution. After creating another conditional, they stalled and tabbed back to ChatGPT. They repeated this process of trying to write on their own, stalling, and going back to ChatGPT several times. The condition, once finished, checked if the number was greater than zero with an else-if checking if it was less than zero. They then placed a while loop below this conditional block. Throughout this time, their gaze moved between code provided by ChatGPT and their own code, but they seemed to be placing the discrete elements from the former into incorrect places in the latter. For instance, when writing the code inside of a while loop, their eyes went to the portion of ChatGPT's code for deciding output. Running this code produced an infinite loop. P17 did not understand how to take the code from ChatGPT and use it to fix their own and did not seem to understand how to put these components together, showing a \mpro \space difficulty.

\begin{figure*}[htbp]
\centering
  \includegraphics[width=1\linewidth]{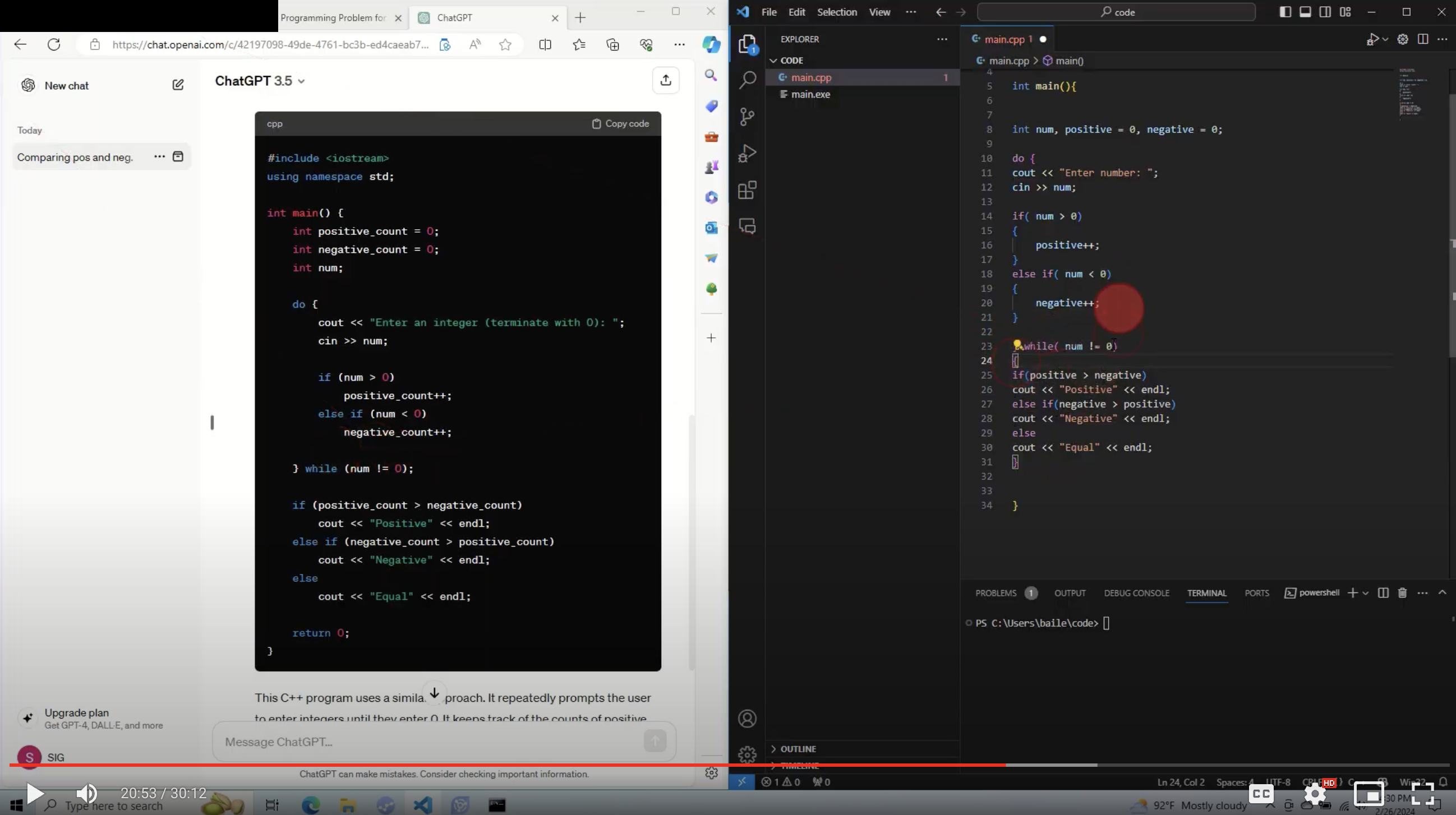}
  \caption{P17 attempting to pattern their code after ChatGPT.}
  \label{fig:P17}
\end{figure*}

After a quick look over their code, P17 went back to ChatGPT, their gaze moving between its code and their code. Unsure how to proceed, they then read the text that ChatGPT included with the code that explained the program. They ran the program again, trying a positive number, then a negative number, and then zero. The first two led to infinite loops while the last ended the program. They then adopted the do-while from ChatGPT, but implemented it incorrectly, which still led to an infinite loop (see Figure \ref{fig:P17}). Finally, they deleted all of the code, copied the code from ChatGPT, and pasted it into the source file. After forgetting to recompile before testing, it still had an infinite loop, which sent them back to ChatGPT and the problem description, trying to find the bug. Unable to find it, they pasted the problem description into ChatGPT again and it provided the same code. They carefully compared the code in the source file to ChatGPT's, but found no discrepancies. So, they re-copied the code from ChatGPT into the source file. This time they recompiled, tested, and succeeded.

When asked about the use of Copilot during the lab session, they said, ``Yes, it makes solving stuff quicker. It doesn't help you, its just autocomplete stuff you would already type.'' Their response about ChatGPT was similarly helpful, saying, ``Yes, I really don't understand code at all, it acts like a teacher to help solve problems.'' Their actions and verbalizations during the lab session did not indicate that Copilot suggested things they would already have in mind to type, nor did they utilize ChatGPT like a teacher.

\subsubsection{P21: \mfor \space and \mmis}
%21 - Treats the program as limited input. Copilot assists.
P21 started by reading the entire problem description and the sample test cases, saying, ``Okay, cool.'' After deleting the warm-up ``Hello World'' code, Copilot generated a long suggestion to solve a different problem. Without even looking at it, they started to declare a variable and then chose instead to write a comment: ``this program shows more negative or positive numbers''. Copilot then generated a line of five variables: num1, num2, etc. They accepted this and deleted the fifth, since the sample test case input was only four numbers before terminating, saying, ``Did they say how many entries? I'm guessing it's just four entries.'' Copilot then suggested two more variables to track positive and negative numbers, which they also accepted. This suggestion by Copilot to have so many variables placed P21 on a path to thinking it was a limited amount of input, rather than an indefinite amount as specified in the problem description. This is evidenced by them next saying, ``I could do it with a while or a for loop, but I guess it's neither.'' They therefore experienced a \mmis \space difficulty.

They then wrote code to accept a number, followed by a conditional to determine which counter to increment. P21 wrote similar code below for each of four input variables until the basic structure was repeated four times. Each time, Copilot accelerated this by suggesting the conditional, which they accepted. They then wrote the conditional logic for outputting which counter was greater. After fixing some minor syntax issues, cleaning up the code, and ensuring spacing was correct for all outputs, they said, ``That looks like that's it! Hope we're good.'' After testing the code locally with the sample test cases, they submitted to Athene, which reported passing 1 of 25 test cases. The test case it passed only had two numbers for input and after reading it they remarked, ``Oh it can accept less than four. We're gonna have to delete all that probably. Gonna need a while loop." 

P21 created a while loop with the condition that the input number be greater than zero. Copilot immediately suggested the correct contents of the loop, which they accepted and then commented-out the four input segments. They then tested the code, got a correct answer, and updated the condition inside the while loop to be ``num = !0''. This did not produce expected behavior, so they changed it to ``num == !0''. This also did not work, so they changed it to ``num < 0 \&\& num > 0''. As they typed this, they said, ``while num is bigger than 0 and less than 0.'' This, of course, would also not work. Since Copilot sent them on the wrong path initially, even though they were now on the correct one, they were too far into the problem solving process to understand what was needed to complete the program. In other words, they hadn't reasoned themselves into this position; Copilot had.

\begin{figure*}[ht]
\centering
  \includegraphics[width=1\linewidth]{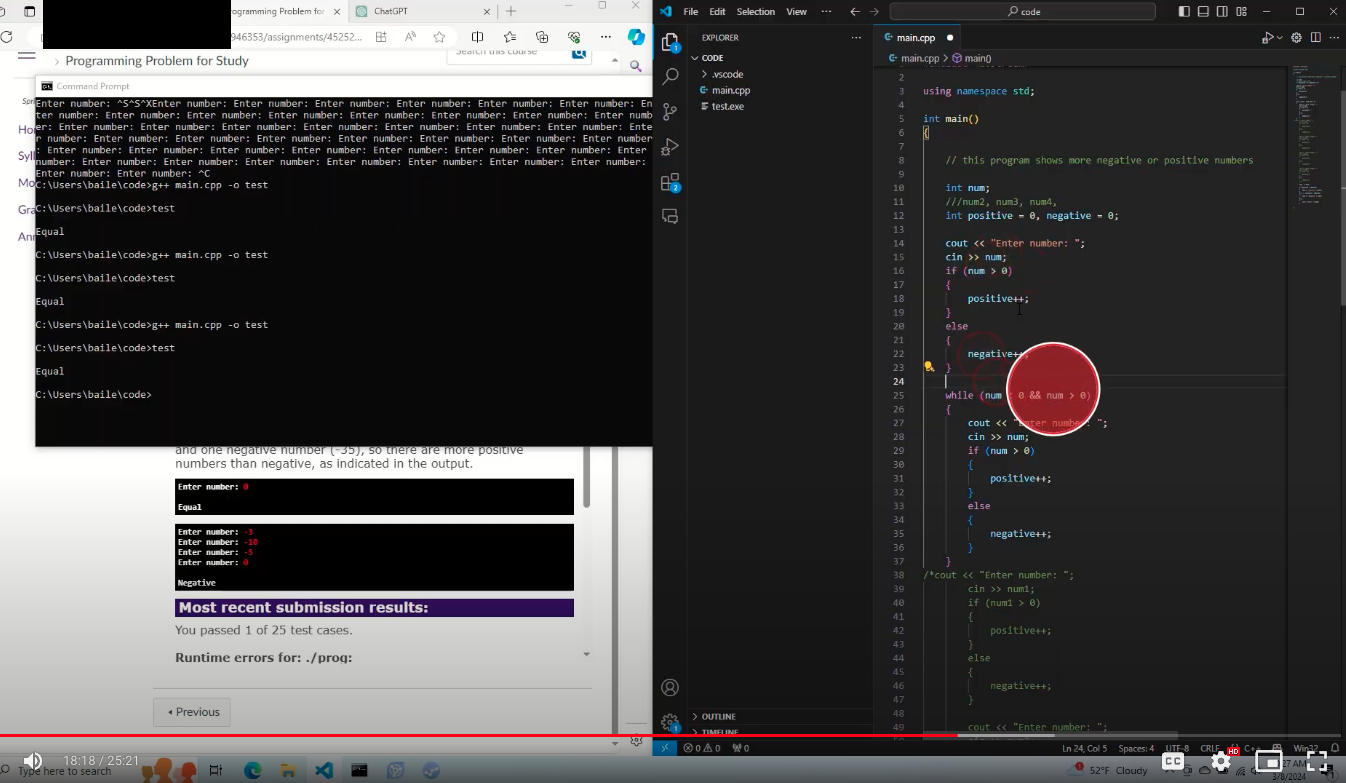}
  \caption{P21 trying to determine why the control flow will not enter the loop.}
  \label{fig:P21}
\end{figure*}

They next added input above the loop and Copilot suggested a condition to count whether the number was positive or negative, which they accepted (see Figure \ref{fig:P21}). Running the code again, it only accepted one number thanks to the condition in the while loop. After looking at the code, they verbalized their current understanding of how to proceed, saying, ``How will I put a while loop inside a while loop as long as it's true? I think I'm spending too much time on that, so I'm just gonna go to ChatGPT.'' They asked it ``what is wrong with my code?'' and pasted the contents of the entire source file into the prompt.

ChatGPT responded with four helpful points to consider to correct the code. However, P21 did not gaze at the explanatory text and instead looked right at the code. They immediately began patterning the code after ChatGPT's, tested it, and then submitted to Athene, correctly passing all test cases.

When asked if they thought Copilot had been helpful during the lab session, they said, ``Yes, it was helpful. It's like you’re having a thought in your mind, and it guides you through it, but sometimes it makes it confusing. Like, you’ll have a vague thought in your mind, then it offers a suggestion, and it just blows it away. So, it just really depends on if you can hold onto the thought in your mind or not.''

%These students had no difficulties:
% 2
% 3
% 5
% 6
% 10
% 12
% 15
% 16
% 18
% 19
% 20

\subsection{Participants Who Accelerated}
Eleven students did not display any metacognitive difficulties. Instead, some used AI in positive ways to help them solve the problem quickly. P3, P19, and P20 each had about a 30\% acceptance rate (see Table \ref{tab.bystudent}), which was much closer to the average acceptance rate for the participants who faced metacognitive difficulties. However, the Copilot suggestions that these three participants utilized were all exactly what they needed. P2, with the lowest acceptance rate of 10\% (2/20), ignored most Copilot suggestions, and is also included below for comparison.

%When asked about Copilot, P3 replied, ``I didn't use it, I try not to.'' Regarding ChatGPT, they said, ``I like to think through the problems on my own.'' Space constraints prevent us from examining these in close detail, but it is clear that some students were able to use GenAI in helpful and ideal ways.

\subsubsection{P20}
P20 ignored multiple Copilot suggestions, reading some and not others, as they focused on writing their solution. It was clear from the eye tracking data that P20 at least glanced at suggestions to determine their utility, saying to one Copilot suggestion, ``No I don’t want that.'' All along the way, they verbalized their ideas and plan for writing the code, maintaining a clear plan throughout. They also accepted multiple Copilot suggestions that completed blocks of code, such as an else after an if. This allowed them to move quickly to a correct solution in just 12 minutes.

%(see Figure \ref{fig:P20}).

%\begin{figure*}[htbp]
%\centering
%  \includegraphics[width=1\linewidth]{Figures/eyetracking/P20-12min-complete.png}
%  \caption{Participant 20 at the end of the session.}
%  \label{fig:P20}
%\end{figure*}

%slow accept!
\subsubsection{P3}
After reading the problem prompt and verbalizing a correct understanding of the problem, P3 started writing comments in the code for scaffolding as they wrote. Although this helped Copilot make informed suggestions, they ignored many of them, only accepting about five one-line suggestions. These accepts were all ``slow accepts'' \cite{prather2024tochi} where the user types out a suggestion character for character. The initial program was completed in 5 minutes, but contained a few scattered syntax errors. For instance, P3 used the keyword ``elif'' instead of ``else if''. However, they eventually realized this error and corrected it without GenAI help, completing the program and passing all test cases in 13 minutes. Interestingly, when asked about Copilot, they replied, ``I didn't use it, I try not to.'' Regarding ChatGPT, they said, ``I like to think through the problems on my own.''

\subsubsection{P19}
P19 read the problem description and immediately started naming their variables, creating: ``num'', ``pos'', and ``neg''. With this information, Copilot immediately suggested a while loop that would do exactly what they wanted, they carefully read it and referenced the problem description, and then accepted it. Below that, by just typing the keyword ``if'', Copilot suggested the entire output control flow, which they also read and then accepted. The output did not quite match what Athene required, so they modified it. They then tested and submitted the code, finishing in just 5 minutes. When asked after the session about whether they thought Copilot had been helpful, they said, ``Definitely sped it up. It gives you a lot of the scaffolding you need to solve the problem so you don't have to worry about some of that.''

\subsubsection{P2}
P2 read the problem description and then said, ``I need a while loop because it needs a sentinel.'' Similar to P19, they immediately created appropriately named variables ``num'', ``pos'', and ``neg''. P2 then started writing code to accept user input and loop through it. Along the way, Copilot suggestions were ignored. Many of these suggestions were misleading (e.g. solving ``more positive or negative'') or incorrect (e.g. output that would not be scored correctly). After completing construction of the loop, P2 started writing the conditionals below. Copilot gave another suggestion for the ''if'' statement and P2 typed through it in a ``slow accept.'' Copilot then correctly suggested the rest of the conditional block (''else if'' and ''else''), which P2 accepted. P2 then compiled and ran the program, fixed an error in their loop, submitted to Athene, and passed all test cases in six minutes.

%=========================
%
%   Discussion
%
%=========================

\section{Discussion}
\subsection{Benefits of GenAI for Novice Programmer Learning}
In answer to RQ1 -- \textit{What benefits do novice programmers receive from using GenAI tools to solve programming problems?} -- we found that several of the participants in this study were able to accelerate to a solution thanks to use of GenAI tools, a finding consistent with prior work \cite{barke2022grounded}. Margulieux et al. examined novice behavior with GenAI tools and found that higher performing students utilized GenAI tools less frequently and later in the problem solving process than those with lower grades and lower self-efficacy \cite{margulieux2024self}. Recent studies examining novice programming behavior with GenAI tools report that some students can use them well, while other students flounder \cite{prather2024tochi}. Other recent studies suggest that while more advanced students may derive disproportionate benefits, those with weaker backgrounds may find the tools less usable and in some cases may be less inclined to trust them or to use them~\cite{hou2024effects, zastudil2023generative}. Similar results are beginning to be discussed in other disciplines \cite{darvishi2024impact}.

One interesting anomaly in the acceleration group is worth exploring. Although P3 claimed they did not use Copilot, we observed them performing two slow accepts. It is possible they felt that they had wanted to do what Copilot was suggesting anyway and so did not feel that this qualified as using Copilot. Given the observed cognitive dissonance reported above in the struggling students, it is also possible that P3 simply was not aware that they were using Copilot in that moment. With its constant stream of little suggestions, GenAI tools like Copilot can fade into the background, making it easy to forget when they are and are not using it. Therefore, it is possible many students may not be fully aware of their GenAI usage and that self-reported data on their usage habits and patterns may not be as reliable as most currently think.

The students in our study who successfully used GenAI seemed better poised to ignore unhelpful and misleading suggestions. It appears that the participants who accelerated were able to quickly recognize ``bad'' GenAI suggestions more effectively than their peers who struggled. In 1994, Marvin Minsky suggested that we often discuss expertise in terms of what people know, but it is just as important to discuss it in terms of their ``negative expertise'' \cite{minsky1994negative}. That is, an expert has seen just as many or more examples of what does \emph{not} work as they have of what \emph{does} work. Minsky writes, ``In order to think effectively, we must `know' a good deal about what not to think! Otherwise we get bad ideas -- and also, take too long.'' \cite{minsky1994negative} From our observations above, it appears that students who were able to successfully utilize GenAI had begun developing this negative expertise in relation to code reading. The ability to ignore incorrect or unhelpful GenAI suggestions appears to be an important skill that should be developed in novices going forward. Mozannar et al. recently showed a possible way to achieve this via reflective retrospective labeling of coding sessions \cite{mozannar2024reading}. Although not in the context of GenAI, Xie et al. proposed using code replays to scaffold novice metacognition \cite{xie2023developing} and future work could replicate their study with GenAI. Finally, future work could explore whether it is good for student learning to expose students to numerous GenAI suggestions through automated assessment to help them determine which ones are helpful and which ones are not.

\subsection{Harms of GenAI to Novice Programmer Learning}
In answer to RQ2 -- \textit{What difficulties do novice programmers face while using GenAI tools to solve programming problems?} -- we noted the same metacognitive difficulties among students consistent with previous findings \cite{prather2018metacognitive}. However, some of the previously identified metacognitive difficulties have been compounded by GenAI. The most common metacognitive difficulty found in our results was \textit{Location}, which GenAI unfortunately facilitated for many of our participants, giving them the illusion of progress. We also identified novel metacognitive difficulties arising from GenAI use: \mpro, \mint, and \mmis \space (see Table \ref{tab.metacognitives}). 
Perhaps most concerning are the ways in which these novel GenAI metacognitive difficulties made it more difficult for participants to become aware of their own lack of understanding. Post-problem interviews confirmed that many of the participants rationalized their usage of GenAI tools in ways that contradicted their words and actions during the sessions.

In the prior study by Prather et al. \cite{prather2018metacognitive}, 11 participants out of 31 did not complete the programming problem within the 35 minute time window. Each one of the students in the prior study realized that they could not solve the problem after spending 35 minutes struggling. However, out of the 10 participants in our study who struggled to complete the problem (out of 21 total participants), 9 utilized GenAI to get to the solution, with only a single struggling student not solving the problem. 
Although they might indeed have solved it on their own within the time limit, our observations of their words and actions lead us to believe they would not have.
From the evidence presented above, it appears that most of these ten who struggled thought they understood more than they actually did. The patterns of behavior above describe how participants were often led along by GenAI such that each step was able to be rationalized as understanding, making it even more difficult for participants to assess their own learning.

Various solutions for issues around metacognition in novice programmers have been proposed and could be insightful here. First, Xie et al. proposed using code replays to scaffold novice metacognition \cite{xie2023developing}. They noticed that pauses in code editing during the replay were fruitful times of reflection for novices and linked this with positive building of self-regulation behaviors. Pauses in code editing with GenAI tools like Copilot are when code suggestions appear, potentially also making this a useful exercise for novices engaged in GenAI-supported coding. These pauses could enable novices to see when they spent too much time reading unhelpful suggestions or when they accepted unhelpful code and reflect on the outcome. Second, Loksa et al. explicitly taught the programming problem solving process to novice programmers, which scaffolded metacognitive skills \cite{loksa2016programming}. They noticed a large and significant increase in self-efficacy in their experimental group relative to their control group. Given that our results linked lower self-efficacy to more metacognitive issues, it seems crucial to explicitly teach the problem solving process and help novice learners reflect on where they are in that process to prevent metacognitive difficulties. Other interventions suggested by Prather et al. \cite{prather2019first} and Pechorina et al. \cite{pechorina2023metacodenition} provide additional fruitful ground for interventions designed to scaffold novice metacognition. Future work should investigate whether these methods are effective interventions to prevent the newly identified metacognitive difficulties presented in this work.

While there is very little work yet on how novice programmers utilize GenAI tools, three recent works illuminate our findings. First, Liu et al.~\cite{liu2024teaching} described implementing a GenAI tool to help students in an introductory programming course. They presented quotes from excited students who described Generative AI as ``like having a personal tutor.'' From the data in our study, P7 self-reported the same feeling and yet we observed that they did not, in fact, use GenAI tools like a personal tutor. In fact, it was quite the opposite. 
%While the student in the class of Liu et al. who said the reported quote may have, in fact, used it like a tutor, it's also possible that many students merely \textit{thought} they used it that way.

Second, recent work by Vadaparty et al. \cite{vadaparty2024cs1llm} describes the usage of GenAI tools from the very beginning in a large introductory programming course. They also reported on survey data from students who took the course and found that students generally liked using GenAI tools and over 50\% of students reported that it positively impacted their learning. However, as the results in our study indicate, there may be more happening here than is at first apparent. Students may feel as though they are learning even as GenAI tools replace critical thinking and problem solving for them. Some of the quotes from Vadaparty et al. seem to get at the rough edges around this problem. One student in the study reported that Copilot helped them understand concepts, but not master them, going faster without fully understanding. Another student said they did not feel confident coding a solution without the help of Copilot. Vadaparty et al. also reported that they anecdotally found students performed worse on code writing exam questions while performing about the same at code tracing and code reading questions. This fits the data from our study that participants were often able to recognize a correct solution, but unable to get there themselves. The theory of programming instruction by Xie et al.~\cite{xie2019theory} states that student learning should start at the code reading and tracing level and then move down into code writing. It appears that for some students, GenAI tools could interfere with this movement into a deeper level of learning programming.

Third, a study in \textit{Communications of the ACM} in March, 2024 by Ziegler et al. investigated how GitHub Copilot is impacting programmer productivity \cite{ziegler2024measuring}. They found that acceptance rate of Copilot's suggestions was highly correlated with perceived productivity. They also report that users with the least experience tend to accept the most suggestions, mirroring similar findings above (see Table \ref{tab.metasnometas}). These data that novices may feel productive by accepting GenAI code suggestions align with our results. However, our data also indicates that this is not a useful way to measure novice programmer interaction with GenAI tools because students could be accepting incorrect code suggestions or suggestions they do not understand. 
Early work on this by Vaithilingam et al. \cite{vaithilingam2022expectation} showed that when asked about what benefits they received from GenAI, many said that it saved them time, despite the fact that the researchers measured no time savings from those using GenAI tools. 

Recent work measuring learning outcomes among introductory programming students using GenAI are in conflict. A study by Xue et al. \cite{xue2024does} reports that ChatGPT has no impact on student learning outcomes in introductory programming. However, Jost et al. found that increased LLM usage negatively correlated with student grades and critical thinking skills \cite{jost2024impact}. Although we did not measure learning outcomes, our results as well as those by Jost et al. show that these matters are far from solved.

%=========================
%
%   Conclusion
%
%=========================
\section{Conclusions}

% The results in our study corroborate and significantly expand on the findings discussed above: 
\textbf{Our findings suggest that students who are already poised to succeed can leverage GenAI to accelerate, while struggling students may be hindered by using GenAI, leaving them with an illusion of competence.}
Participants with higher grades were less likely to have GenAI metacognitive difficulties and participants with lower self-efficacy were more likely to have metacognitive difficulties.  
More insidiously, the nuances of \textit{how} students struggle with GenAI may not be immediately apparent from an overall view of grades in a course. Although our sample size was too small to tell if prior programming experience, socio-economic background, race, or gender are factors in GenAI harming novice programmer experience, the data we do have warrants urgent future exploration into these open questions. 

Ultimately, our findings confirm that the most commonly used GenAI tools are not informed by pedagogy.  ChatGPT is a general purpose tool that students are increasingly using for their work~\cite{prather2023robots, liu2024teaching, vadaparty2024cs1llm, hou2024effects}. GitHub Copilot is becoming more commonly used by students as well~\cite{becker2023programming, prather2023robots, hou2024effects, sheard2024instructor}, but it is targeted at professional developers. These two tools are situated on the edges of the \textit{broad} to \textit{domain-specific} spectrum. Perhaps it is because of this that many students lack the ability to helpfully engage with either of these tools. We also cannot reasonably expect students to abstain from using these tools since many are freely available and banning them does not seem like a viable strategy \cite{lau2023ban}. 

Therefore, it is imperative that we scaffold the novice programmer experience of GenAI tools from the very beginning by lowering the metacognitive demand on the user \cite{tankelevitch2024metacognitive}. There are already multiple novice-friendly tools that attempt to do this by exposing students to GenAI in scaffolded ways, such as \textit{Prompt Problems} \cite{denny2024prompt}, \textit{CodeHelp} \cite{liffiton2024codehelp}, CodeAid \cite{kazemitabaar2024codeaid}, and \textit{Ivie} \cite{yan2024ivie}. Some ``digital TAs'' are also emerging that scaffold students through the problem solving process without providing answers \cite{denny2024desirable}. All of these tools can show students how to use GenAI responsibly without replacing the critical thinking component of the programming problem solving process. Another key component for moving forward in the era of GenAI is to finally invest in explicitly teaching novice programmers metacognitive behaviors and skills \cite{margulieux2024self, tankelevitch2024metacognitive, xie2023developing}. Our findings above illustrate well that a failure in metacognition can lead to poor results with using GenAI. Without addressing these issues, the gap between the well-prepared and the under-prepared in computing education may grow even wider.

\begin{acks}
This research is funded by the Google Award for Inclusion Research Program and was also partially supported by the Research Council of Finland (Academy Research Fellow grant number 356114).
% Juho's grant? added
\end{acks}

\bibliographystyle{ACM-Reference-Format}
\bibliography{ref-ai, ref-metacog}

%\appendix

\end{document}